\documentclass{article}
\usepackage[final]{neurips_data_2022}  %

\usepackage{wrapfig}
\usepackage{natbib}
\usepackage[utf8]{inputenc} 
\usepackage[T1]{fontenc}    
\usepackage{hyperref}       
\usepackage{url}            
\usepackage{booktabs}       
\usepackage{amsfonts}       
\usepackage{nicefrac}       
\usepackage{microtype}      
\usepackage[dvipsnames]{xcolor}         
\usepackage{amsmath}
\usepackage{amssymb}
\usepackage{bm}
\usepackage{relsize}
\usepackage{pgfplots}
\usepackage[utf8]{inputenc}
\usepackage{pgfplots}
\DeclareUnicodeCharacter{2212}{−}
\usepgfplotslibrary{groupplots,dateplot}
\usetikzlibrary{patterns,shapes.arrows}
\pgfplotsset{compat=newest}
\usepackage{xcolor}
\usepackage{booktabs,multirow,array}
\usepackage{longtable}
\usepackage[font=small,labelfont=bf]{caption}
\usepackage{subcaption}
\usepackage{float}
\usepackage{algpseudocode,algorithm,algorithmicx}
\usepackage{arydshln}
\usepackage{setspace}
\usepackage{ctable}
\usepackage{tikz}
\usetikzlibrary{shapes.geometric, arrows}
\usepackage{longtable}
\usepackage{enumitem}  %

\renewcommand{\vec}[1]{{\bm #1}}

\newcommand{\x}{\vec{x}}
\newcommand{\y}{\vec{y}}
\newcommand{\z}{\vec{z}}
\renewcommand{\u}{\vec{u}}

\newcommand{\vtheta}{\vec{\theta}}
\newcommand{\vlambda}{\vec{\lambda}}

\newcommand{\f}{f_\text{aug}}

\title{Myriad: a real-world testbed to bridge \\
trajectory optimization and deep learning}

\author{%
  Nikolaus H. R. Howe \\
  Mila, Université de Montréal\\
  \texttt{niki.howe@mila.quebec} \\
  \And Simon Dufort-Labbé\\
  Mila, Université de Montréal\\
  \And Nitarshan Rajkumar \\
  University of Cambridge\thanks{Work done while at Mila,
  Université de Montréal.} \\
  \And Pierre-Luc Bacon \\
  Mila, Université de Montréal, Facebook CIFAR AI, IVADO\\
}

\begin{document}

\maketitle

\begin{abstract}
We present Myriad, a testbed written in JAX which enables machine learning researchers to benchmark imitation learning and reinforcement learning algorithms against trajectory optimization-based methods in real-world environments.
Myriad contains 17 optimal control problems presented in continuous time which span medicine, ecology, epidemiology, and engineering.
As such, Myriad strives to serve as a stepping stone towards application of modern machine learning techniques for impactful real-world tasks.
The repository also provides machine learning practitioners access to trajectory optimization techniques, not only for standalone use, but also for integration within a typical automatic differentiation workflow.
Indeed, the combination of classical control theory and deep learning in a fully GPU-compatible package unlocks potential for new algorithms to arise.
We present one such novel approach for use in optimal control tasks.
Trained in a fully end-to-end fashion, our model leverages an implicit planning module over neural ordinary differential equations, enabling simultaneous learning and planning with unknown environment dynamics.
All environments, optimizers and tools are available in the software package at \url{https://github.com/nikihowe/myriad}.
\end{abstract}

\section{Introduction}\label{sec:intro}
The rapid progress of machine learning (ML) algorithms is made clear by the yearly improvement we see on standard  ML benchmarks \citep{imagenet, mujoco, ale}.
Inevitably, the popularity of a given testbed creates a positive feedback effect, encouraging researchers to develop algorithms that achieve good performance on that set of tasks \citep{relevant, deeprlthatmatters}.
We believe it is crucial that our algorithms be well-suited for  positive-impact, real-world applications. As such, we must be able to train and test them on real-world-relevant tasks.

\begin{wrapfigure}{r}{0.5\textwidth}
    \centering
    \vspace{-10pt}
    \resizebox{0.5\textwidth}{!}{
    \input{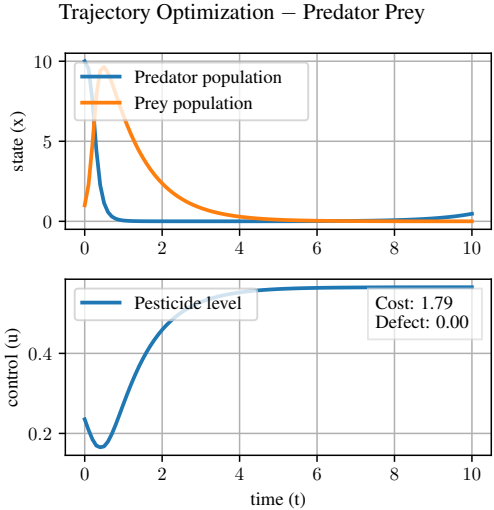}
    }
    \caption[Predator Prey Trajectory Optimization]{
    The optimal trajectory of pesticide use over time and resulting population dynamics in the Predator Prey domain, which is included in Myriad (see Table~\ref{tab:envs} for a complete list of environments).
    Direct single shooting (see Section~\ref{sec:traj_opt}) was used to compute the optimal control trajectory.
    }
    \vspace{10pt}
    \label{fig:predator_prey_traj_opt}
    \vspace{-0.5cm}
\end{wrapfigure}

\setcounter{footnote}{0}
To this end, we present Myriad, a real-world testbed for optimal control methods such as imitation learning and reinforcement learning (RL).
Myriad differs from previous testbeds in several key aspects.
First and most importantly, all tasks are inspired by real-world problems, with applications in medicine, ecology, epidemiology, and engineering.
Second, Myriad is, to our knowledge, the first repository that enables deep learning methods to be combined seamlessly with traditional trajectory optimization techniques.
Figure \ref{fig:predator_prey_traj_opt} shows a visualization of applying a one such trajectory optimization technique on an environment implemented in Myriad.
Third, the system dynamics in Myriad are continuous in time and space, offering several advantages over discretized environments.
On the one hand, these algorithms are adaptable to changing sampling frequencies, or even irregularly spaced data.
At the same time, using continuous-time dynamics gives the user freedom to choose between integration techniques, and opens the door to efficient variable-step integration methods. These can take advantage of the local environment dynamics to effectively trade off speed and accuracy of integration \citep{rk45}.

While the field of control theory has yielded practical approaches to solve a plethora of problems in the industrial setting \citep{lenhart, betts, chemical}, adoption of such methods within the ML community has been limited.
This is in part due to the historical focus of ML on scalability, and of control theory on robustness and optimality, and is further exacerbated by the lack of tools to run control algorithms in a deep learning setting.
Yet trajectory optimization techniques can offer excellent performance in systems with known dynamics, and thus serve as a solid benchmark against which to test RL techniques.
Additionally, trajectory optimization offers several advantages over standard RL, such as the ability to impose safety constraints, which is crucial in real-world applications \citep{betts, chemical}.
As such, the Myriad repository allows for the combination of trajectory optimization and deep learning techniques into powerful hybrid algorithms.
As an example, we implement an end-to-end trained implicit planning imitation learning algorithm, and benchmark its performance alongside that of trajectory optimization on known and learned dynamics models.

The following sections present various aspects of Myriad, starting with a review of related work in \textbf{Section~\ref{sec:related_work}}.
In turn, \textbf{Section~\ref{sec:myriad}} gives an overview of the Myriad repository and describes several of the available control environments.
The subsequent sections can be thought of as both presenting the tools in Myriad, as well as describing the building blocks used to create the aforementioned imitation learning algorithm.
To start, \textbf{Section~\ref{sec:traj_opt}} describes direct single shooting, a standard trajectory optimization technique, and
\textbf{Section~\ref{sec:scale}} shows how we can leverage GPU-accelerated first-order methods to use trajectory optimization in a machine learning setting.
\textbf{Section~\ref{sec:sysid}} presents the system identification problem, and how Myriad can be used to learn neural ordinary differential equation models \citep{node} of unknown system dynamics.
Finally, \textbf{Section~\ref{sec:end-to-end}} presents a new deep imitation learning algorithm which includes a control-oriented inductive bias, trained end-to-end using tools from Myriad.
\textbf{Section~\ref{sec:conclu}} concludes and discusses the limitations and potential societal impact of this work.

We summarize our main contributions as follows:
\begin{itemize}[noitemsep,nolistsep]
    \item We present a \textbf{testbed for real-world tasks}, including learning dynamics models from data and the problem of optimal control. The testbed contains 17 continuous-time real-world tasks, and it is straight-forward to add additional systems to the repository.
    \item We provide a set of plug-and-play \textbf{differentiable trajectory optimization algorithms} implemented in JAX \citep{jax}, which can be used standalone or in conjunction with machine learning techniques.
    \item We introduce a \textbf{novel control-oriented imitation learning algorithm} which combines optimal control with deep learning. The tools in Myriad enable us to develop this method and compare its performance to control methods which leverage known or learned dynamics.
    \item We collect \textbf{benchmark reference scores} for most environments, achieved using classical optimal control techniques with the true system dynamics, as well as benchmark scores achieved using trajectory optimization on two kinds of learned dynamics model. See Appendix~\ref{app:benchmark_scores} for these scores along with details on how they were obtained.
\end{itemize}

\section{Related work}
\label{sec:related_work}

\textbf{Testbed}:
Within the context of RL, several testbeds have been influential in pushing forward the state-of-the art, notably OpenAI Gym \citep{gym} and the Arcade Learning Environment \citep{ale}.
Yet, these environments are inherently discrete in time, and focus primarily on game-like settings, abstracting away much of the challenge when working with real-world problems.
There also exists a rich collection of software for robotics tasks, some of which are differentiable \citep{drake, mujoco, pybullet}.
However, these are narrow in scope, only focusing on physics simulation, which makes them challenging to leverage to build a testbed for other kinds of real-world problems.

While there exist ML testbeds for real-world tasks \citep{wilds2021}, Myriad is to our knowledge the first to focus on learning and control, and to leverage trajectory optimization in a deep learning context. Indeed, even packages which provide real-world trajectory optimization problems often rely on symbolic differentiation and non-differentiable solvers to compute optimal trajectories \citep{beluga, casadi}, making them unsuitable for use in a deep learning workflow.

\textbf{Algorithm}:
Various attempts have been made to include optimal control techniques within a larger neural network architecture, for example by creating a differentiable sub-network \citep{ondifferentiating} with gradients computed via loop unrolling \citep{okada, pereira} or implicit differentiation \citep{mairal, optnet, diffmpc, pontryagin_diff_programming}.
Yet, these works tend to focus on specific settings:
\citet{pereira} and \citet{okada} apply model predictive control to DAGGER \citep{dagger} and to a Monte-Carlo-based recurrent setting respectively, while
\citet{mairal} and \citet{optnet} instead focus on specific problem formulations, such as solving a lasso \citep{lasso} problem or a quadratic program.

The works which are algorithmically most similar to ours are \citet{diffmpc} and \citet{pontryagin_diff_programming}, both of which attempt to learn arbitrary dynamics in an end-to-end fashion, with some caveats.
Most crucially, the techniques of both works can only learn the parameters of dynamics (and cost) functions \emph{of which the functional form is already known}.
This relies on a human expert to craft a sufficiently accurate description of the dynamics, which for complex dynamical systems can be an insurmountable task.
Myriad overcomes this challenge by using neural ordinary differential equations (Neural ODEs) \citep{node} to learn arbitrary system dynamics directly from data.
Additionally, both works only consider problems that are discrete in time, and also rely on non-differentiable solvers and thus can only compute gradients at convergence via the implicit function theorem.

Furthermore, \citet{diffmpc} leverage a quadratic program solver to work with local quadratic approximations of the dynamics, as opposed to fully nonlinear dynamics.
As such, their approach is unsuitable for use with learned dynamics parametrized by neural networks.
In contrast, the solvers in Myriad can be applied in nonconvex settings, unlocking the use of neural network-based dynamics models.

While the approach of \citet{pontryagin_diff_programming} can be used with arbitrary dynamics, the authors treat system identification and optimal control in isolation of one another, while Myriad enables the user to close the loop by using gradients from control to improve system identification, and by gathering data based on the current best guess for controls.
Additionally, \citet{pontryagin_diff_programming} rely on a neural policy network to compute controls in the optimal control setting, while Myriad instead plans controls directly from learned dynamics.

To our knowledge, while there is active and diverse research in the areas related to this work, there has remained an ongoing lack of real-world environments and plug-and-play trajectory optimization tools for a deep learning workflow. 
As such, we believe that Myriad, which is implemented entirely in JAX \citep{jax}, is well-placed to bridge trajectory optimization and deep learning with its collection of real-world tasks and differentiable optimal control algorithms.

\section{Myriad: environments and optimizers}
\label{sec:myriad}
Myriad was developed with the machine learning community in mind, and offers environments spanning medicine, ecology, epidemiology, and engineering (a full list is provided in Table~\ref{tab:envs}).
The repository contains implementations of various trajectory optimization techniques, including single and multiple shooting \citep{betts}, trapezoidal and Hermite-Simpson collocation \citep{kelly}, and the indirect forward-backward sweep method \citep{lenhart}.
We also offer algorithms for learning dynamics, both to identify the unknown parameters of an a priori model, and to learn a black-box Neural ODE model \citep{node}.

With the exception of off-the shelf nonlinear program solvers such as \texttt{ipopt} \citep{ipopt} and \texttt{SLSQP} \citep{SLSQP}, every aspect of the systems and trajectory optimizers is differentiable, allowing flexible use and easy incorporation in a deep learning practitioner's workflow.

Myriad is extensible, enabling straightforward addition of new environments, trajectory optimization techniques, nonlinear program solvers, and integration methods to the repository:

\begin{itemize}

\item
We consider \emph{control environments} (which we also call \emph{systems} or \emph{environments}) specified by their dynamics function, cost function, start state, and final time. A system can optionally include a required terminal state, a terminal cost function, and bounds on the state and controls.
In order to create a new system, the user can extend  \texttt{FiniteHorizonControlSystem}, an abstract class defined in \texttt{systems/base.py}.
We present some of these environments below; for a table of all environments and link to documentation including full environment descriptions, see Appendices~\ref{app:repo_and_documentation}~and~\ref{app:environments}.

\item
A \emph{trajectory optimizer} has an objective function, a constraint function, control and state bounds, an initial decision variable guess, and an  unravel function for sorting the decision variable array into states and controls.
To implement a trajectory optimizer, the user can extend the abstract class \texttt{TrajectoryOptimizer}, an abstract class defined in \texttt{optimizers/base.py}.
A table of the trajectory optimizers currently available in Myriad is given in Appendix~\ref{app:optimizers_and_integration_methods}.

\item
A \emph{nonlinear program solver} is set to have the same function signature as those used by standard off-the-shelf solvers such as \texttt{ipopt} and \texttt{SLSQP}.
To implement a new solver, the user can create a function with the same signature as those in \texttt{nlp\_solvers/}. 
\end{itemize}

Below we give an overview of some of the environments in Myriad.
For a full description of the environments, see the repository documentation; a link is in Appendix~\ref{app:repo_and_documentation}.
We note that many of these environments were inspired by work of \citet{lenhart}, and both the formulation and framing of such environments should be attributed to them.

\textbf{Medicine and Epidemiology}
\begin{itemize}
\item In \textbf{Cancer Treatment}, we want to determine the optimal administration of chemotherapeutic drugs to reduce the number of tumour cells.
Control is taken to be the strength of the drug, while the cost functional is the normalized tumour density plus the drug side effects, as done by \citet{fister}.
The dynamics assume that tumour cells will be killed in proportion to the tumour population size \citep{skipper}.
\item In \textbf{Epidemic}, we aim to find the optimal vaccination strategy for managing an epidemic.
Control is the percentage rate of vaccination (what proportion of the population is newly vaccinated every day), while the cost functional is the number of infectious people plus a cost quadratic in vaccination effort.
The dynamics follow a SEIR model \citep{seir}.
\item In \textbf{Glucose}, we want to regulate blood glucose levels in someone with diabetes, following the approach of \citet{eisen}.
Control is set to be insulin injection level, and the cost is quadratic in both the difference between current and optimal glucose level, as well as in the amount of insulin used.
\end{itemize}

\textbf{Ecology and science}

\begin{itemize}
\item In the \textbf{Bear Populations} setting, we manage the metapopulation of bear populations in a forest and a national park within the forest, an important problem when it comes to ensuring species preservation while also avoiding bears in human-populated areas, based on the work of \citet{bears}.
The controls are the rates of hunting in the forest and the national park.
The cost is the number of bears that exit the forest, plus a hunting cost in each location.
\item In \textbf{Mould Fungicide}, we want to decrease the size of a mould population with a fungicide.
Control is the amount of fungicide used, while the cost is quadratic in both population size and amount of fungicide used. 
\item In \textbf{Predator Prey}, we wish to decrease the size of a pest population by means of a pesticide, which acts as control.
We assume that the pest population is prey to a predator in the ecosystem, which we do not wish to impact.
The dynamics follow a Lotka-Volterra model, and the cost is the final prey population, plus a quadratic control cost.
\end{itemize}

\subsubsection*{Control}
We also include several classical control environments, such as \textbf{Pendulum} (the dynamics and cost of which match OpenAI Gym \citep{gym}), \textbf{Cart-Pole Swing-Up} as presented by \citet{kelly}, and \textbf{Mountain Car}, which also matches Gym \citep{gym} except for the function describing the hill, which was changed from sinusoidal to quadratic to improve stability during Neural ODE-based system identification. While these are standard problems, we believe it worthwhile to reproduce them for study in the trajectory optimization setting. We also include other control problems such as the challenging \textbf{Rocket Landing} domain described by \citet{rocket_landing}, and the forced \textbf{Van der Pol} oscillator, as presented by \citet{casadi}.

\section{Trajectory optimization}
\label{sec:traj_opt}

Many control problems can be formulated in the language of trajectory optimization, in which an optimization technique is used to find a control trajectory which minimizes an integrated cost.
While trajectory optimization approaches are rarely considered by RL practitioners, they often provide good solutions when using a known dynamics model, and thus can serve as a useful benchmark for many optimal control tasks.
To give a flavour of the techniques used in trajectory optimization, here we present the standard method of direct single shooting \citep{betts}, which is implemented in Myriad alongside other algorithms.

Letting $\u$ and $\x$ represent control and state functions,  $c$ the instantaneous cost and $f$ the system dynamics, the trajectory optimization problem can be written as
\begingroup
\allowdisplaybreaks
\begin{equation}
\begin{aligned}
    \min_{\u(t) \; \forall t \in [t_s, t_f]} \quad & \int_{t_s}^{t_f} c(\x(t), \u(t), t) \; dt\\
    \text{such that} \quad & \dot \x(t) = f(\x(t), \u(t)) \; \forall t \in [t_s, t_f] \\
    \text{with} \quad & \x(t_s) = \x_s \\
    \text{and*} \quad & \x(t_f) = \x_f \\
    \text{and*} \quad & \x_\text{lower}(t) \leq \x(t) \leq \x_\text{upper}(t) \; \forall t \in [t_s, t_f] \\
    \text{and*} \quad & \u_\text{lower}(t) \leq \u(t) \leq \u_\text{upper}(t) \; \forall t \in [t_s, t_f]
\end{aligned}
\end{equation}
\endgroup
where asterisks indicate optional constraints.
Note that we allow time-dependent cost, but assume time-independent dynamics.
First, we augment the system dynamics with the instantaneous cost:
\begin{align}
    \f(\x(t), \u(t), t) = \begin{bmatrix} f(\x(t), \u(t)) \\ c(\x(t), \u(t), t) \end{bmatrix}.
\end{align}
Then the integral
\begin{equation}
    \begin{bmatrix} \x_s \\ 0 \end{bmatrix} + \int_{t_s}^{t_f} \f(\x(t), \u(t), t) \; dt = \begin{bmatrix} \x_f \\ c_f \end{bmatrix}
\end{equation}
will contain the integrated cost -- the objective we want to minimize -- as its final entry.
Let $\psi$ be a function which, given a sequence of controls and a timestamp, returns an interpolated control value.\footnote{How this interpolation is performed depends on the integration method applied. Matching the control discretization with a fixed integration timestep circumvents the need for explicit interpolation.}

Letting $\x(t_s) = \x_s$ and $c(t_s) = 0$, we can construct 
the following nonlinear program (NLP):
\begingroup
\allowdisplaybreaks
\begin{equation}
\begin{aligned}
    \text{decision variables} \quad & \hat \u_0, \hat \u_1, \hat \u_2, \ldots, \hat \u_N \\
    \text{objective} \quad & \left[ \int_{t_s}^{t_f} \f \left(\begin{bmatrix} \x(t) \\ c(t) \end{bmatrix}, \psi(\hat \u_{0:N}, t), t \right) \; dt \right] \texttt{[-1]} \\
    \text{equality constraints*} \quad & \x_f = \x_s + \int_{t_s}^{t_f} f(\x(t), \psi(\hat \u_{0:N}, t)) \; dt \\ 
    \text{inequality constraints*} \quad & \u^\text{lower}_i \leq \hat \u_i \leq \u^\text{upper}_i \quad \text{for } i = 0, \ldots, N \\
\end{aligned}
\label{eq:ss}
\end{equation}
\endgroup

\begin{wrapfigure}{r}{0.5\textwidth}
    \centering
    \vspace{0.35cm}
    \resizebox{0.5\textwidth}{!}{
    \input{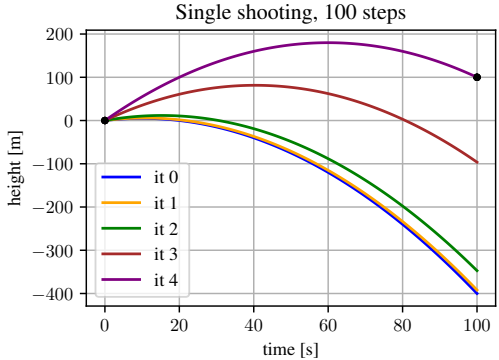}
    }
    \vspace{-10pt}
    \caption[]{
    Trajectories computed after 0 to 4 iterations of direct single shooting.
    At each iteration, the gradient is propagated through the forward integration back to the initial parameters, which are updated to decrease the final defect.
    }
    \label{fig:ss}
\end{wrapfigure}

To gain more intuition about direct single  shooting, we visualize a toy problem of projectile motion, in which we are trying to get a projectile to an altitude of 100m after exactly 100s by choosing a launch velocity.
Under simplifying assumptions, given state $\x = [ x, \dot x ]^\top$, the dynamics can be written as $f(\x) = [ \dot x, -g ]^\top$, where $g$ is gravitational acceleration.
Figure \ref{fig:ss} shows the outcome of applying direct single shooting to this problem.

Direct single shooting is the perhaps the simplest of the trajectory optimization techniques, but it comes with several shortcomings, of which we mention the two most impactful.
First, direct single shooting does not allow us to impose constraints on the state trajectory found by the trajectory optimization solver.
Yet, for an algorithm to be safe to apply in real-world settings, it is crucial that the user be able to restrict the system to a set of safe states (in robotics, avoiding collisions; in chemical engineering, avoiding unsafe pressure/temperature levels, etc.).
Second, direct single shooting is inherently sequential, making it possibly slower and less effective in long-horizon tasks due to integration time and vanishing gradients.
Other optimization techniques, such as direct multiple shooting \citep{betts} and direct collocation \citep{kelly} support parallelization, and might be much more efficient when solving over long time horizons.
For a table of the trajectory optimization techniques provided in Myriad (which includes those mentioned above), see Appendix~\ref{app:optimizers_and_integration_methods}.

\section{Constrained optimization at scale}
\label{sec:scale}

The nonlinear programs of Section~\ref{sec:traj_opt} are usually solved using second-order techniques based on Newton's method \citep{sqp, nocedal}.
We would like to be able to solve these nonlinear programs at scale, leveraging the advantages that GPU-based optimization has brought to deep learning \citep{alexnet, tensorflow}.
Unfortunately, many software implementations
such as \texttt{ipopt} and \texttt{SLSQP} are restricted to CPU, instead relying on sparse matrix operations for computational efficiency \citep{ipopt, SLSQP}.
While effective for small models, such higher-order techniques struggle when applied over a large number of parameters, due to the size of the resulting Hessian matrix \citep{kfac}.
Indeed, since GPU computation is more suitable for dense matrix operations \citep{gpu}, the emphasis on sparsity in traditional solvers is of little help when it comes to deep learning applications. 
Other problems further exacerbate the challenge of using these solvers for ML: not only do higher-order techniques tend to perform poorly in high-dimensional settings due to the prevalence of saddle-points \citep{saddle_point}; these solvers are also non-differentiable, making them practically impossible to use in methods where we need to propagate gradients through the solve itself (such as the imitation learning technique presented in Section~\ref{sec:end-to-end}).

Here we present a simple technique with which we have found success on many Myriad environments, and which runs fully on GPU.
Let $f$ be the objective function, and $h$ the equality constraints (for example, these could be the objective and equality constraints from
Eq.~\eqref{eq:ss} or Eq.~\eqref{eq:ms}).
We use $\y = [\x, \u]^\top$ to denote decision variables of the NLP.
The Lagrangian of our NLP is then
\begin{equation}
    \mathcal{L}(\y, \vec \lambda) = 
f(\y) + \vec \lambda^\top h(\y), \label{eq:lag1}
\end{equation}
where $\vlambda$ are known as the Lagrange multipliers of the problem.
We see that a solution to the NLP will correspond to the solution of the min-max game: $\min_\y \max_\vlambda \mathcal{L}(\y, \vec \lambda)$ \citep{kushnerandyin}.
In particular, this solution will satisfy the first-order optimality condition that
$(D_1 \; \mathcal{L})(\y^\star, \vlambda^\star) = 0$ \citep{bertsekas}.
We can attempt to find a solution by applying a first-order Lagrangian method \citep{arrow, uzawa} to find $(\y^\star, \vlambda^\star)$:
\begin{equation}
\begin{aligned}
    \y^{(i+1)} &\gets \y^{(i)} - \eta_\y \cdot (D_1 \; f)(\y^{(i)}, \vlambda^{(i)}) \\
    \vlambda^{(i+1)} &\gets \vlambda^{(i)} + \eta_\vlambda \cdot (D_2 \; f)(\y^{(i)}, \vlambda^{(i)}).
\end{aligned}
\end{equation}

As an instance of gradient descent-ascent \citep{gda},
this method can suffer from oscillatory and even divergent
dynamics \citep{polyak}. One way to mitigate this
is the extragradient method \citep{extragradient, gauthier}.
Instead of following the gradient at the current
iterate, extragradient performs a ``lookahead step'', effectively
evaluating the gradient that would occur at a future step.
It then applies the lookahead gradient to the current iterate.
\begin{equation}
\begin{aligned}
    \bar \y^{(i)} &\gets \y^{(i)} - \eta_\y \cdot (D_1 \; f)(\y^{(i)}, \vlambda^{(i)}) \\
    \bar \vlambda^{(i)} &\gets \vlambda^{(i)} + \eta_\vlambda \cdot (D_2 \; f)(\y^{(i)}, \vlambda^{(i)}) \\
    \y^{(i+1)} &\gets \y^{(i)} - \eta_\y \cdot (D_1 \; f)(\bar \y^{(i)}, \bar \vlambda^{(i)}) \\
    \vlambda^{(i+1)} &\gets \vlambda^{(i)} + \eta_\vlambda \cdot (D_2 \; f)(\bar \y^{(i)}, \bar \vlambda^{(i)}).
\end{aligned}
\end{equation}
This approach has seen recent success in
the generative adversarial model literature,
and it seems likely that further
improvements can be made by leveraging
synergies with game-theoretic
optimization \citep{workshop1, dragan, GANstable}.

In practice, since we are considering real-world systems, we often want to restrict the trajectories the agent can take through state space to a safe subset.
There are several ways to include inequalities when using a Lagrangian-based approach; they are described in Appendix~\ref{app:inequality_constraints}.

\section{System identification}

Sections \ref{sec:traj_opt} and \ref{sec:scale} showed how to solve a standard trajectory optimization problem assuming known dynamics.
While such techniques can be used as a basic benchmark for RL algorithms, it is often more realistic to compare an RL approach with a setting in which trajectory optimization is performed on \emph{learned} dynamics.
To this end, we turn our attention to learning system dynamics from data, i.e., the problem of \emph{system identification} (SysID) \citep{sysid}.

In control theory, SysID is typically performed to learn the parameters of a highly structured model developed by field experts.
Indeed, such highly structured models have been used even in recent work at the intersection of learning and control \citep{diffmpc, pontryagin_diff_programming}.
Not only is this task comparatively simple due to having to learn only a handful of  parameters; in the case of identifiable systems, it is also easy to verify the accuracy of the learned model by simply checking the values of the learned parameters.

Yet the construction of a structured model relies on the ability of a human expert to accurately describe the dynamics, which is a lengthy process at best, and impossible for sufficiently complex systems. RL circumvents this issue either by not using a world model, or by building one from data \citep{sutton, mbrl}.
In order to provide a trajectory optimization benchmark, we must learn a model directly from data. We do this by modelling the dynamics of a system with a Neural ODE \citep{node}: a natural fit when it comes to continuous systems.
While Neural ODEs have not yet been extensively studied in the context of controllable environments \citep{controlled, dynode}, it is not challenging to extend them to this setting.
In this case we would like to find Neural ODE parameters $\vtheta$ which best approximate the true dynamics:
\begin{align}
    f(\x(t), \u(t), \vtheta) 
    \equiv \texttt{apply\_net}\left(\vtheta,
        [ \x(t), \u(t) ]^\top\right)
    \approx f(\x(t), \u(t)),
\end{align}
where $f(\x(t), \u(t))$ is the true dynamics function.
In order to train
this model, consider a trajectory of
states\footnote{Myriad offers several methods for
generating trajectory datasets,
including uniformly at random, Gaussian random walk,
and sampling around a candidate control trajectory.}
$\x$, sampled with noise from the true
dynamics,
given controls $\u_{0:N}$.
We would like our model to 
predict this trajectory. In particular,
$\tilde{\x}$ should approximate
$\x$:

\begin{equation}
    \tilde \x =
    \left[
        \x_0, \x_0 + \int_{t_0}^{t_1} f(\x(t), \psi(\u_{0:N}, t), \vtheta) \; dt,
        \ldots, \x_0 + \int_{t_0}^{t_N} f(\x(t), \psi(\u_{0:N}, t), \vtheta) \; dt \\
    \right].
\end{equation}

We minimize the mean squared error between the two trajectories ($N$ is number of timesteps, $D$ is state dimension, giving $\x$ and $\tilde \x$ dimensions $(D, N)$).
The loss is then calculated as\footnote{In practice, the loss calculation is performed in parallel over a minibatch of training trajectories.}:
\begin{align}
    L(\hat \vtheta) &= \frac{1}{ND}
    \| \tilde \x - \x \|^2_E,
    \label{eq:loss}
\end{align}
where $\| \bullet \|^2_E$ is the squared Euclidean norm (sum of squares of elements).

\label{sec:sysid}
\begin{wrapfigure}{r}{0.5\textwidth}
    \centering
    \vspace{-0.5cm}
    \resizebox{0.5\textwidth}{!}{\input{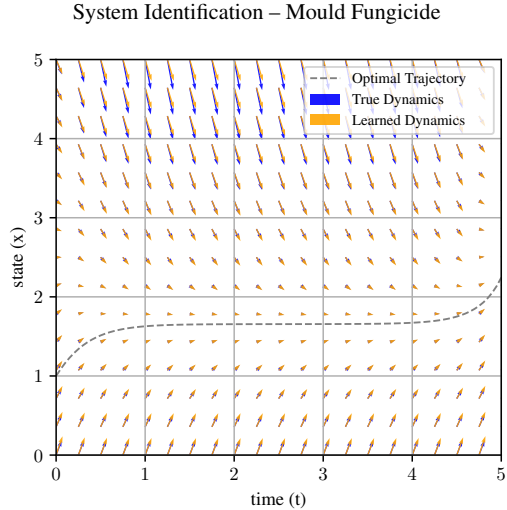}}
    \caption[Dynamics study on Neural ODE Mould Fungicide model]{Comparison of true dynamics and learned dynamics (Neural ODE model)
    when applying optimal controls in the Mould Fungicide domain.
    We observe that the dynamics learned via SysID closely match the true dynamics for this problem.
    }
    \vspace{-0.5cm}
    \label{fig:mf_mle_vector}
    \vspace{0.2cm}
\end{wrapfigure}

\section{End-to-end SysID and control}
\label{sec:end-to-end}

In the Neural ODE setting, the individual parameters no longer convey an intuitive physical meaning. Yet, we can still compare the learned to the true dynamics by considering the effect of a given control sequence across a range of states.
An example of such visualization is shown in Figure \ref{fig:mf_mle_vector}, which compares the Neural ODE learned model's dynamics with those of the true dynamics on a mould fungicide domain \citep{lenhart}.
We use automatic differentiation to calculate the gradient of the loss in Eq.\ \ref{eq:loss} with respect to network parameters; another approach is to apply the adjoint sensitivity method as done by \citet{node}.

As in other areas of machine learning, RL has seen increasing interest in forgoing the use of explicit models, instead structuring the policy to include a planning inductive bias such that an agent can perform \emph{implicit} planning \citep{vit, xlvin, diffmpc, pontryagin_diff_programming}.
A classic example is value iteration networks \citep{vit}, which replace explicit value iteration with an inductive bias in the form of a convolutional neural network \citep{fukushima, cnn}.

Inspired by implicit planning, we consider a fully differentiable algorithm which performs trajectory optimization on an implicit model.
By propagating gradients through the trajectory optimization procedure itself, the agent can learn directly from the loss received from acting in the real environment.
In order to describe this approach -- a form of ``unrolled optimization'' \citep{unrolled} -- we consider a modification to the Lagrangian of Eq.\ \eqref{eq:lag1}, adding parameters $\vtheta$ which parametrize the underlying dynamics function.
We let $\y$ represent the primal variables (control and state decision variables), $\vlambda$ the dual variables (Lagrange multipliers), $f$ the objective function, and $h$ the equality constraints.
To simplify notation, we let $\z = [\y, \vlambda]^\top$, which gives the Lagrangian:
\begin{equation}
    \mathcal{L}\left(\vtheta, \z \right) = f(\y, \vtheta) + \vlambda^\top h(\y, \vtheta).
    \label{eq:lag2}
\end{equation}
Let $\psi$ be a function representing the  nonlinear program solver, which takes  parameter values $\hat \vtheta$ and returns $\hat \z$, and let $L$ be the loss function of Eq.~\ref{eq:loss}.
We would like to  propagate the gradient of $L$ with respect to $\hat \vtheta$ through $\psi$ at our current decision variables $\hat \y$.
The basic procedure to achieve this is shown in Algorithm \ref{alg:basic_algorithm}.

    \begin{algorithm}[H]
      \setstretch{1.05}
      \caption{End-to-End $-$ Theory}
        \label{alg:basic_algorithm}
      \begin{algorithmic}[1]
        \State Initialize $\hat \u_{0:N}$, $\hat \vtheta$ with random values
        \While{$\hat \u_{0:N}$, $\hat \vtheta$ not converged}
          \State $\hat \z \gets \psi(\hat \vtheta)$ \Comment{solve NLP represented by Eq.\ \eqref{eq:lag2}}
          \State $\hat \x, \hat \u_{0:N}, \hat \vlambda \gets \hat \z$ \Comment{extract controls}
          \State $\hat \vtheta \gets$ update using $(D \; (L \circ \psi))(\hat \vtheta)$ \label{line:gd}
        \EndWhile
        \State \Return $\hat \u_{0:N}$
      \end{algorithmic}
    \end{algorithm}

The clear challenge is the implementation of Line \ref{line:gd}.
By the chain rule we have that
\begin{equation}
    \left(D \; (L \circ \psi) \right)(\vtheta) = (D \; L)(\psi(\vtheta)) \cdot (D \; \psi)(\vtheta).
\end{equation}
The first term, $(D \; L)(\psi(\vtheta))$, can simply be calculated using automatic differentiation in the imitation learning setting, or using a gradient approximation method in the RL setting \citep{reinforce}.
The calculation of $(D \; \psi)(\vtheta)$ is more challenging, since it involves differentiating through the NLP solver.
A natural first approach is to apply the implicit function theorem (IFT), which suggests that for $(\vtheta, \z)$ such that $\z = \psi(\vtheta)$ and  $(D_1 \; \mathcal{L})(\vtheta, \z)$ is near zero, we have
\begin{equation}
    (D \; \psi)(\vtheta) = -\left(D_2 D_1 \; \mathcal{L} \right)^{-1}(\vtheta, \z) \cdot \left( D_1^2 \; \mathcal{L} \right)(\vtheta, \z).
\end{equation}
In practice, we experienced several drawbacks when using this method.
Most notably, we found the requirement that $(D_1\; \mathcal{L})(\vtheta, \z)$ be near zero in order for the implicit function theorem to hold particularly challenging, since an unreasonable amount of computation must be spent to achieve such high accuracy from the NLP solver.

A practical workaround is to use a \emph{partial} solution at each timestep, and take gradients through an unrolled differentiable NLP solver.
By performing several gradient updates per iteration and warm-starting each step at the previous iterate, we are able to progress towards an optimal solution with a computationally feasible approach.
We reset the warm-start after a large number of iterations, as in \citep{pontryagin_diff_programming}, to avoid catastrophic forgetting of previously-seen dynamics.
This approach, which we use in our imitation learning algorithm implementation, is presented in Algorithm \ref{alg:practical_e2e}.

\begin{algorithm}[H]
  \setstretch{1.05}
  \caption{End-to-End Approach $-$ Practice}
    \label{alg:practical_e2e}
  \begin{algorithmic}[1]
    \State Initialize $\hat \u_{0:N}, \hat \vtheta$ with random values
    \While{$\hat \u_{0:N}, \hat \vtheta$ not converged}
      \State $\hat \z, \texttt{dz\_dtheta} \gets$ simultaneously take several steps of $\psi(\hat \vtheta)$
      and accumulate gradients
      \State $\hat \x, \hat \u_{0:N}, \hat \vlambda \gets \hat \z$ \Comment{extract controls}
      \State $\texttt{dL\_dz} \gets (D \; L)(\hat \z)$
      \Comment using automatic differentiation or gradient approximation
      \State $\texttt{dL\_dtheta} \gets \texttt{dL\_dz} \cdot \texttt{dz\_dtheta} $
      \Comment{apply the chain rule}
      \State $\hat \vtheta \gets$ update with $\texttt{dL\_dtheta}$
    \EndWhile
    \State \Return $\hat \u_{0:N}$
  \end{algorithmic}
\end{algorithm}

\begin{figure}
    \centering
    \vspace{-1.25cm}
    \resizebox{0.75\textwidth}{!}{
    \includegraphics{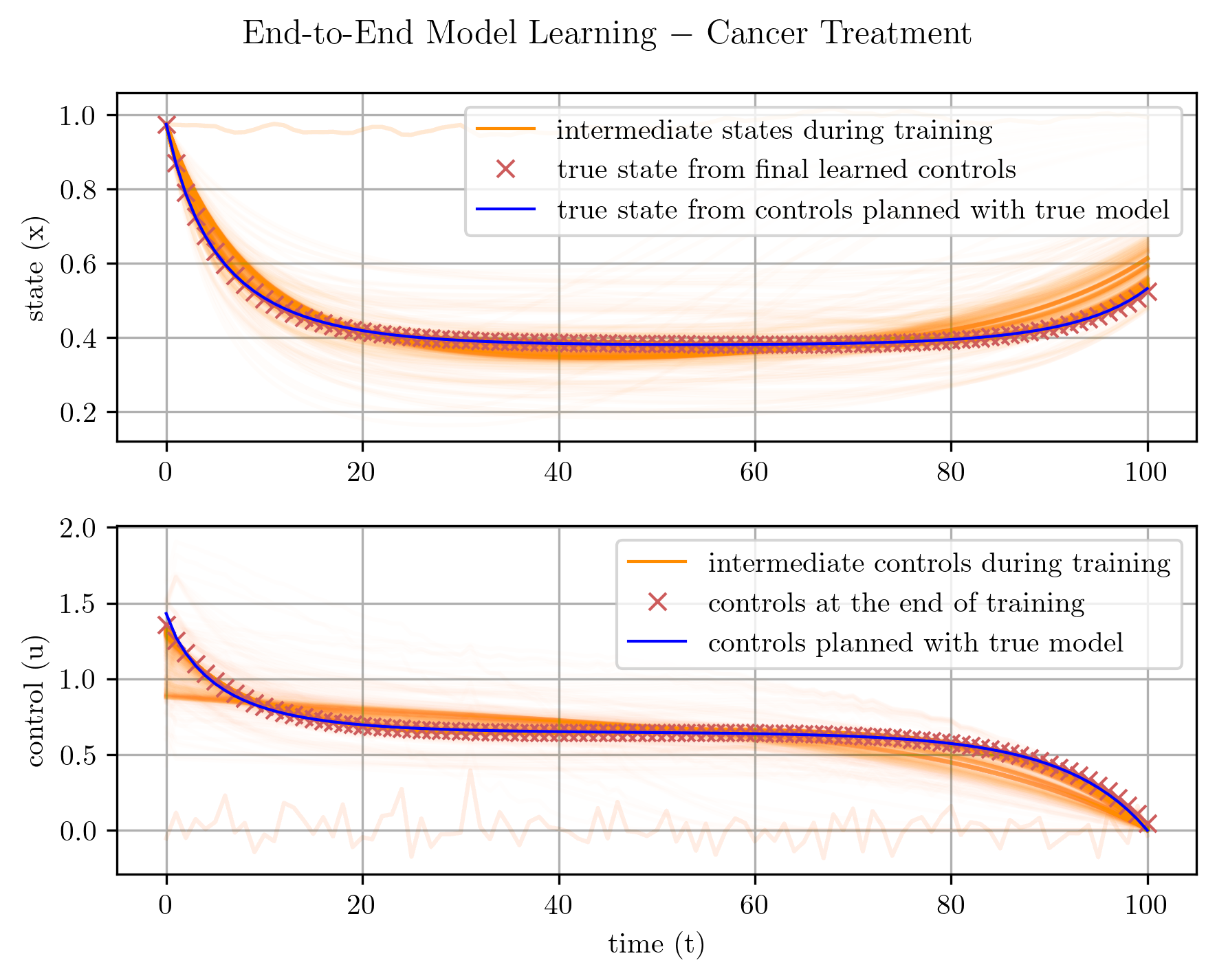}
    }
    \caption[Visualization of end-to-end Neural ODE model 
    learning on Cancer Treatment domain]{
    Visualization of how the controls, and corresponding states, evolve over  the course of training a Neural ODE model end-to-end on the Cancer Treatment domain.
    The control trajectory, and corresponding state trajectory, are sampled regularly over the course of training.
    Each is plotted with a low alpha value to show where the learning procedure spent time during training.
    }
    \vspace{-0.75cm}
    \label{fig:ct_node_e2e_cool_plot}
    \vspace{0.25cm}
\end{figure}

We find that Algorithm \ref{alg:practical_e2e} is able to learn effective models and propose good controls for several environments.
To gain intuition about how the model learns its environment over time, we take snapshots of the controls proposed by our algorithm over the course of training.
We give an example of this in Figure \ref{fig:ct_node_e2e_cool_plot}, which shows the progress of end-to-end training of a Neural ODE model on a cancer treatment domain \citep{lenhart}.

\section{Conclusion}
\label{sec:conclu}

Implemented in JAX, the systems and tools in Myriad fit seamlessly in a deep learning workflow, and can serve both to develop new algorithms and benchmark them against existing optimal control techniques.
The current environments span medicine, ecology, epidemiology, and engineering, and special attention has been made to allow easy integration of new environments, optimizers, and nonlinear programming tools.
We showcase the power of Myriad's tools by developing a novel control-oriented imitation learning algorithm which combines optimal control with deep learning in an end-to-end trainable approach.
Not only does the algorithm achieve good performance on several environments; Myriad also enables comparison of this new technique with traditional trajectory optimization over fixed or learned system dynamics.

\textbf{Limitations:} There are several limitations to the Myriad repository as well as to the imitation learning algorithm developed with Myriad tools.
First and foremost is the fact that many of the environments in Myriad were selected in part because they were known to be amenable to solution via traditional optimal control methods.
As a result, the dimensionality of state observations in all Myriad environments is low (<10 dimensions) compared with pixel-based tasks which require representation learning of visual features \citep{human_level}.
Thus, Myriad is not at present useful for benchmarking the effectiveness of learned visual representations in a deep RL setting.
Technical limitations are also present: for now, only fixed-step integration methods are supported, limiting our ability to take advantage of regions of simple dynamics for faster integration. Another beneficial enhancement would be the implementation of variable scaling, which would help avoid numerical stability issues which can sometimes occur when integrating through rapidly changing dynamics.
Finally, an important shortcoming of the imitation learning algorithm is its inability to learn a cost function, which in the general setting is not known during learning.
It would be desirable to expand the model's learning capability to include a cost function -- similar to the approach of \citet{pontryagin_diff_programming} but using a more expressive Neural ODE model \citep{node} -- and benchmark its performance compared with the current version.

\textbf{Societal impact:} Many of the environments presented in Myriad are inspired by real-world problems \citep{lenhart, betts}. 
However, we caution that they should not by themselves be used for medical, ecological, epidemiological, or any other real-world application, since they abstract away real-world application-specific details which must be examined and approached on a case-by-case basis by experts in the domain. 
For example, it is important to consider the safe limits of operation and have a fallback control routine in the case of controlling robots or industrial processes.
Our goal is that the Myriad testbed will help build interest within the machine learning community to bring our algorithms to application in impactful real-world settings.

\textbf{Acknowledgements:} Thank you to Lama Saouma for input regarding  the feasibility of an end-to-end approach for SysID and Control and to Andrei M.\ Romascanu for feedback on a previous version.
The authors are also thank Hydro-Québec, Samsung Electronics Co., Ldt., Facebook CIFAR AI, and IVADO for their funding, and Calcul Québec and Compute Canada for compute resources.

\bibliographystyle{apalike}
\def\bibname{Bibliography}
\bibliography{ref}

\clearpage
\section*{Checklist}

\begin{enumerate}

\item For all authors...
\begin{enumerate}
  \item Do the main claims made in the abstract and introduction accurately reflect the paper's contributions and scope?
    \answerYes{See Section~\ref{sec:intro} describing fully the paper structure.}
  \item Did you describe the limitations of your work?
    \answerYes{See Section~\ref{sec:conclu}.}
  \item Did you discuss any potential negative societal impacts of your work?
    \answerYes{See Section~\ref{sec:conclu}.}
  \item Have you read the ethics review guidelines and ensured that your paper conforms to them?
    \answerYes{}
\end{enumerate}

\item If you are including theoretical results...
\begin{enumerate}
  \item Did you state the full set of assumptions of all theoretical results?
    \answerNA{}
	\item Did you include complete proofs of all theoretical results?
    \answerNA{}
\end{enumerate}

\item If you ran experiments (e.g. for benchmarks)...
\begin{enumerate}
  \item Did you include the code, data, and instructions needed to reproduce the main experimental results (either in the supplemental material or as a URL)?
    \answerYes{See Abstract for URL, as well as Appendix~\ref{app:repo_and_documentation}}.
  \item Did you specify all the training details (e.g., data splits, hyperparameters, how they were chosen)?
    \answerYes{See Appendix~\ref{app:benchmark_scores} for training details of the benchmark references scores.}
	\item Did you report error bars (e.g., with respect to the random seed after running experiments multiple times)?
    \answerNA{The testbed allows for generation of new benchmark scores; we reported the best performance we were able to achieve for a given setting.}
	\item Did you include the total amount of compute and the type of resources used (e.g., type of GPUs, internal cluster, or cloud provider)?
    \answerNA{The testbed allows for generation of new benchmark scores; the reported scores can be generated on a personal laptop.}
\end{enumerate}

\item If you are using existing assets (e.g., code, data, models) or curating/releasing new assets...
\begin{enumerate}
  \item If your work uses existing assets, did you cite the creators?
    \answerYes{All adapted/inspired content was thoroughly cited.}
  \item Did you mention the license of the assets?
    \answerYes{The license is mentioned in the repository.}
  \item Did you include any new assets either in the supplemental material or as a URL?
    \answerYes{See last line of the abstract.}
  \item Did you discuss whether and how consent was obtained from people whose data you're using/curating?
    \answerNA{}
  \item Did you discuss whether the data you are using/curating contains personally identifiable information or offensive content?
    \answerNA{}
\end{enumerate}

\item If you used crowdsourcing or conducted research with human subjects...
\begin{enumerate}
  \item Did you include the full text of instructions given to participants and screenshots, if applicable?
    \answerNA{}
  \item Did you describe any potential participant risks, with links to Institutional Review Board (IRB) approvals, if applicable?
    \answerNA{}
  \item Did you include the estimated hourly wage paid to participants and the total amount spent on participant compensation?
    \answerNA{}
\end{enumerate}

\end{enumerate}

\clearpage
\appendix
\section{Repository and documentation}
\label{app:repo_and_documentation}

The repository for this project can be found at
\url{https://github.com/nikihowe/myriad}.

The documentation for this project can be found at
\url{https://nikihowe.github.io/myriad/html/myriad}.

\section{Implemented in Myriad}
\label{app:implemented_in_myriad}

Here we give scores for 17 real-world environments, as well as for a ``simple case'' environment, which is a toy setting to be used for initial algorithm testing.
\subsection{Environments}
\label{app:environments}
\begin{table}[H]
\centering
\begin{tabular}{@{}llll@{}}
\toprule
Name               & Brief Description                           & Fixed $\x_T$ & Terminal Cost \\ 
\specialrule{.3em}{.2em}{.2em}
Bacteria*           & Manage bacteria population levels           & No                   & Yes           \\
\hline\noalign{\vskip 0.5ex}
Bear Populations*   & Manage metapopulation of bears              & No                   & No            \\
\hline\noalign{\vskip 0.5ex}
Bioreactor*         & Grow bacteria population                    & No                   & No            \\
\hline\noalign{\vskip 0.5ex}
Cancer Treatment*   & Decrease tumour size via chemotherapy       & No                   & No            \\
\hline\noalign{\vskip 0.5ex}
Cart-Pole Swing-Up & Swing up pendulum by translating pivot       & Yes                  & No            \\
\hline\noalign{\vskip 0.5ex}
Epidemic*           & Control epidemic via vaccination            & No                   & No            \\
\hline\noalign{\vskip 0.5ex}
Glucose*            & Manage blood glucose via insulin injections & No                   & No            \\
\hline\noalign{\vskip 0.5ex}
Harvest*            & Maximize harvest yield                      & No                   & No            \\
\hline\noalign{\vskip 0.5ex}
HIV Treatment*      & Manage HIV via chemotherapy                 & No                   & No            \\
\hline\noalign{\vskip 0.5ex}
Mould Fungicide*    & Control mould population via fungicide      & No                   & No            \\
\hline\noalign{\vskip 0.5ex}
Mountain Car       & Drive up valley with limited force          & Yes                  & No            \\
\hline\noalign{\vskip 0.5ex}
Pendulum           & Swing up pendulum by rotating pivot         & Yes                  & No            \\
\hline\noalign{\vskip 0.5ex}
Predator Prey*      & Minimize pest population                    & Yes                  & Yes           \\
\hline\noalign{\vskip 0.5ex}
Rocket Landing     & Land a rocket                               & Yes                  & No            \\
\hline\noalign{\vskip 0.5ex}
Simple Case        & Use for initial algorithm testing            & No                  & No            \\
\hline\noalign{\vskip 0.5ex}
Timber Harvest*     & Optimize tree harvesting                    & No                   & No            \\
\hline\noalign{\vskip 0.5ex}
Tumour*             & Block tumour blood supply                   & No                   & Yes           \\
\hline\noalign{\vskip 0.5ex}
Van Der Pol        & Forced Van der Pol oscillator               & Yes                  & No            \\ \bottomrule
\\[-5pt]
\end{tabular}
\caption{The environments currently available in the Myriad repository. 
Environments with an (*) were inspired by corresponding examples described by \citet{lenhart}.
Some environments require that the system end up in a specific state at the end of an episode.
Also, some environments impose a final cost in addition to the instantaneous cost, calculated based on the system's final state.
For a detailed description and motivation of each environment, see the documentation linked to in the Myriad repository.}
\label{tab:envs}
\end{table}

\clearpage

\subsection{Optimizers and integration methods}
\label{app:optimizers_and_integration_methods}

\begin{table}[H]
\centering
\begin{tabular}{@{}llll@{}}
\toprule
Method Name                 & Direct / Indirect & Sequential / Parallel & Integration Method    \\ 
\specialrule{.3em}{.2em}{.2em}
Single Shooting             & Direct            & Sequential      & Any                   \\
\hline\noalign{\vskip 0.5ex}
Multiple Shooting           & Direct            & Partially Parallel    & Any                   \\
\hline\noalign{\vskip 0.5ex}
Trapezoidal Collocation     & Direct            & Parallel        & Trapezoidal Rule      \\
\hline\noalign{\vskip 0.5ex}
Hermite-Simpson Collocation & Direct            & Parallel        & Simpson's Rule        \\
\hline\noalign{\vskip 0.5ex}
Forward-Backward Sweep      & Indirect          & Sequential      & Runge-Kutta 4th Order \\ \bottomrule
\\[-5pt]
\end{tabular}
\caption{The trajectory optimization techniques available in the Myriad repository.
Direct techniques discretize the control problem and solve the resulting nonlinear program.
Indirect methods first augment the problem with an adjoint equation, before discretizing and solving.}
\label{tab:opts}
\end{table}

\begin{table}[H]
\centering
\begin{tabular}{@{}lll@{}}
\toprule
Integration Method    & Explicit / Implicit & Gradient Evaluations \\ 
\specialrule{.3em}{.2em}{.2em}
Euler                 & Explicit            & 1                    \\
\hline\noalign{\vskip 0.5ex}
Heun                  & Explicit            & 2                    \\
\hline\noalign{\vskip 0.5ex}
Midpoint              & Explicit            & 2                    \\
\hline\noalign{\vskip 0.5ex}
Runge-Kutta 4th Order & Explicit            & 4                    \\
\hline\noalign{\vskip 0.5ex}
Trapezoidal           & Implicit            & NA                   \\
\hline\noalign{\vskip 0.5ex}
Simpson               & Implicit            & NA                   \\ \bottomrule
\\[-5pt]
\end{tabular}
\caption{The integration methods available in the Myriad
repository.}
\label{tab:ints}
\end{table}

\clearpage
\section{Benchmark scores}
\label{app:benchmark_scores}

We provide tables of benchmark scores for several different algorithms.

\subsection{Trajectory optimization on true dynamics}
To begin with, in Table~\ref{tab:benchmark} we show the scores achieved by a trajectory optimization algorithm on all 17 real-world environments, as well as for the ``simple case'' environment.
The cost is the integrated cost over the time horizon of the problem, plus any terminal cost.
The defect is the difference between the final state and the desired final state of the system.
To reproduce these results, choose a system in \texttt{config.py} and uncomment \texttt{run\_trajectory\_opt} in \texttt{run.py}.
Note that for this setting, no training occurs: we simply run trajectory optimization on the true dynamics model.

\setlength\LTleft{0pt}
\setlength\LTright{0pt}
\begin{table}[H]
\centering
\begin{tabular}{@{}l p{6cm} l l@{}}
\toprule
System & \begin{tabular}{@{}l} Parameters \end{tabular} & \begin{tabular}{@{}l} Cost \end{tabular} & \begin{tabular}{@{}l} Defect \end{tabular} \\
\specialrule{.3em}{.2em}{.2em}
\nopagebreak Bacteria & \raggedright $A$:~1, $B$:~1, $C$:~1 & $-7.98$ & NA \\[2pt]
\hline\noalign{\vskip 0.5ex}
\nopagebreak Bear Populations & \raggedright $r$: 0.1, $K$:~0.75, $m_f$:~0.5, $m_p$:~0.5 & $12.28$ & NA \\[2pt]
\hline\noalign{\vskip 0.5ex}
\nopagebreak Bioreactor & \raggedright $G$: 1, $D$: 1 & $-1.39$ & NA \\[2pt]
\hline\noalign{\vskip 0.5ex}
\nopagebreak Cancer Treatment & \raggedright $\delta$: 0.45, $r$: 0.3 & 20.57 & NA \\[2pt]
\hline\noalign{\vskip 0.5ex}
\nopagebreak Cart-Pole Swing-Up & \raggedright $g$: 9.81, $m_1$: 1, $m_2$:~0.3, $\ell$:~0.5 & 87.78 & $\begin{bmatrix} 0 & 0 & 0 & 0 \end{bmatrix}^\top$ \\[2pt]
\hline\noalign{\vskip 0.5ex}
\nopagebreak Epidemic & \raggedright \text{see repository} & 13.40 & NA \\[2pt]
\hline\noalign{\vskip 0.5ex}
\nopagebreak Glucose & \raggedright $a$: 1, $b$: 1, $c$: 1 & 1354.02 & NA \\[2pt]
\hline\noalign{\vskip 0.5ex}
\nopagebreak Harvest & \raggedright $A$:~5, $k$:~10, $m$:~0.2, $M$:~1 & $-6.51$ & NA \\[2pt]
\hline\noalign{\vskip 0.5ex}
\nopagebreak HIV Treatment & \raggedright $s$:~10, $m_1$:~0.02, $m_2$:~0.5, $m_3$:~4.4, $r$:~0.03, $T_\text{max}$:~1500, $k$:~0.000024, $N$:~300, $A$:~0.05 & $-823.13$ & NA \\[2pt]
\hline\noalign{\vskip 0.5ex}
\nopagebreak Mould Fungicide & \raggedright $r$: 0.3, $M$: 10 & 23.50 & NA \\[2pt]
\hline\noalign{\vskip 0.5ex}
\nopagebreak Mountain Car & \raggedright $g$:~0.0025, $p$:~0.0015 & 8.57 & $\begin{bmatrix}0 & 0\end{bmatrix}^\top$ \\[5pt]
\hline\noalign{\vskip 0.5ex}
\nopagebreak Pendulum & \raggedright $g$:~10, $m$:~1, $\ell$:~1 & 25.75 & $\begin{bmatrix}0 & 0\end{bmatrix}^\top$ \\[2pt]
\hline\noalign{\vskip 0.5ex}
\nopagebreak Predator Prey & \raggedright $d_1$:~0.1, $d_2$:~0.1 & 1.79 & 0 \\[2pt]
\hline\noalign{\vskip 0.5ex}
\nopagebreak Rocket Landing & \raggedright $g$:~9.81, $m$:~100000, $\ell$:~50, $w$:~10 & 178.02 & $\begin{bmatrix} -172.78 \\ 0 \\ -1413.87 \\ -195.49 \\ -36.04 \\ -2.28 \end{bmatrix}$ \\[29pt]
\hline\noalign{\vskip 0.5ex}
\nopagebreak Simple Case & \raggedright $A$:~1, $B$:~1, $C$:~4 & $-1.35$ & NA \\[2pt] 
\hline\noalign{\vskip 0.5ex}
\nopagebreak Timber Harvest & \raggedright $K$: 1 & $-5104.67$ & NA \\[2pt] 
\hline\noalign{\vskip 0.5ex}
\nopagebreak Tumour & \raggedright $\xi$:~0.084, $b$:~5.85, $d$:~0.00873, $G$:~0.15, $\mu$:~0.02 & 7571.67 & NA \\[2pt]
\hline\noalign{\vskip 0.5ex}
\nopagebreak Van Der Pol & \raggedright $a$: 1 & 2.87 & $\begin{bmatrix} 0 & 0 \end{bmatrix}^\top$ \\
\bottomrule \\
\end{tabular}
\caption{
Summary of performance of direct single shooting on the various environments. For these experiments, we used one shooting trajectory with 100 controls. The Heun method was used for integration. We used \texttt{ipopt} to solve the resulting nonlinear program.}
\label{tab:benchmark}
\end{table}

\clearpage

\subsection{Trajectory optimization on learned dynamics}

In addition to performance results using trajectory optimization on the true dynamics models, we also provide benchmark scores of trajectory optimization on models which have been learned from data.

\subsubsection{Parametric models}
Table~\ref{tab:mle_parametric} shows the results of performing optimization on structured parametric models (known dynamics, unknown coefficients) of which the parameters are learned from data, for 12 of the 17 real-world environments, and for the ``simple case'' environment.
For convenience, the corresponding performance achieved using trajectory optimization on the true model is also provided (to the left of the ``/'').
To reproduce these results, choose a system in \texttt{config.py} and uncomment \texttt{run\_mle\_sysid} in \texttt{run.py}.
Run with the default hyperparameters in \texttt{config.py}, which were chosen via a small amount of trial-and-error.
In this setting, we found that a much smaller data regime (train set of 10 trajectories instead of 100) also led to good performance, indicating that using a structured model significantly simplifies the system identification problem.

\setlength\LTleft{0pt}
\setlength\LTright{0pt}
\begin{table}[H]
\centering
\begin{tabular}{@{}l p{4cm} l l@{}}
\toprule
System & \begin{tabular}{@{}l} Parameters \\ (true/learned) \end{tabular} & \begin{tabular}{@{}l} Cost \\ (best/achieved) \end{tabular} & \begin{tabular}{@{}l} Defect \\ (best/achieved) \end{tabular} \\
\specialrule{.3em}{.2em}{.2em}
\nopagebreak Bacteria & \raggedright $A$:~1/1, $B$:~1/1, $C$:~1/1 & $-7.98/{-7.98}$ & NA \\[2pt]
\hline\noalign{\vskip 0.5ex}
\nopagebreak Bear Populations & \raggedright $r$: 0.1/0.1, $K$:~0.75/0.75, $m_f$:~0.5/0.5, $m_p$:~0.5/0.5 & $12.28/12.28$ & NA \\[2pt]
\hline\noalign{\vskip 0.5ex}
\nopagebreak Bioreactor & \raggedright $G$: 1/1, $D$: 1/1 & $-1.39/{-1.39}$ & NA \\[2pt]
\hline\noalign{\vskip 0.5ex}
\nopagebreak Cancer Treatment & \raggedright $\delta$: 0.45/0.45, $r$: 0.3/0.3 & 20.57/20.57 & NA \\[2pt]
\hline\noalign{\vskip 0.5ex}
\nopagebreak Cart-Pole Swing-Up & \raggedright $g$: 9.81/9.81, $m_1$: 1/1, $m_2$:~0.3/0.3, $\ell$: 0.5/0.5 & 87.78/87.78 & $\begin{bmatrix} 0/0 \\ 0/0 \\ 0/0 \\ 0/0 \end{bmatrix}$ \\[17pt]
\hline\noalign{\vskip 0.5ex}
\nopagebreak Glucose & \raggedright $a$: 1/1, $b$: 1/1, $c$: 1/1 & 1354.02/1354.02 & NA \\[2pt]
\hline\noalign{\vskip 0.5ex}
\nopagebreak Mould Fungicide & \raggedright $r$: 0.3/0.3, $M$: 10/10 & 23.50/23.50 & NA \\[2pt]
\hline\noalign{\vskip 0.5ex}
\nopagebreak Mountain Car & \raggedright $g$:~0.0025/0.0025, $p$:~0.0015/0.0015 & 8.57/8.57 & $\begin{bmatrix}0/0 & 0/0\end{bmatrix}^\top$ \\[5pt]
\hline\noalign{\vskip 0.5ex}
\nopagebreak Pendulum & \raggedright $g$:~10/11.943, $m$:~1/0.701, $\ell$:~1/1.194 & 25.75/25.49 & $\begin{bmatrix}0/0.04 & 0/0.04\end{bmatrix}^\top$ \\[2pt]
\hline\noalign{\vskip 0.5ex}
\nopagebreak Predator Prey & \raggedright $d_1$:~0.1/0.1, $d_2$:~0.1/0.1 & 1.79/1.79 & 0/0 \\[2pt]
\hline\noalign{\vskip 0.5ex}
\nopagebreak Timber Harvest & \raggedright $K$: 1/1 & $-5104.67/{-5104.67}$ & NA \\[2pt] 
\hline\noalign{\vskip 0.5ex}
\nopagebreak Tumour & \raggedright $\xi$:~0.084/0.084, $b$:~5.85/5.170, $d$:~0.00873/0.00873, $G$:~0.15/0.15, $\mu$:~{0.02/{$-0.660$}} & 7571.67/7571.73 & NA \\[2pt]
\hline\noalign{\vskip 0.5ex}
\nopagebreak Van Der Pol & \raggedright $a$: 1/1 & 2.87/2.87 & $\begin{bmatrix} 0/0 & 0/0 \end{bmatrix}^\top$ \\
\bottomrule \\
\end{tabular}
\caption[Learned model performance across variety of systems]{
Summary of performance of trajectory optimization on parametric models with parameters learned from data on a variety of environments.
The first column lists the environment, followed by the true and learned parameters, and then the cost (and if applicable, defect) resulting from those parameters.}
\label{tab:mle_parametric}
\end{table}

\clearpage

\subsubsection{Neural ODE models}

In general when acting in real-world environments, we do not know the system dynamics -- or even a parametric model describing the general form of the system dynamics -- beforehand.
As such, it is desirable to be able to learn system dynamics entirely from data.
Table~\ref{tab:node_mle_sysid} shows the performance of trajectory optimization on a Neural ODE model learned from data, for 12 of the 17 real-world environments, and for the ``simple case'' environment.
To reproduce these results, choose a system in \texttt{config.py} and uncomment \texttt{run\_node\_mle\_sysid} in \texttt{run.py}.
Use the default hyperparameters, except as indicated in the final two columns of Table~\ref{tab:node_mle_sysid}, which show the number of trajectories per dataset and total number of datasets used during training.
To select these hyperparameters, we ran training with train sets of size 10 and 100 with up to 5 datasets trained on sequentially (except in Cart-Pole Swing-Up, where we used up to 10 datasets).
We then chose the dataset size and number of datasets which led to the best performance when performing trajectory optimization on the learned model.

\setlength\LTleft{0pt}
\setlength\LTright{0pt}
\begin{table}[H]
\centering
\begin{tabular}{@{}l l l l l@{}}
\toprule
System & {\centering \begin{tabular}{@{}l} Cost \\ (best/achieved) \end{tabular}} & {\centering \begin{tabular}{@{}l} Defect \\ (best/achieved) \end{tabular}} & \# Traj.s/D.set & \# D.sets \\
\specialrule{.3em}{.2em}{.2em}
\nopagebreak Bacteria & $-7.98/{-2.72}$ & NA & 10 & 5 \\[2pt]
\hline\noalign{\vskip 0.5ex}
\nopagebreak Bear Populations & $12.28/12.29$ & NA & 10 & 4 \\[2pt]
\hline\noalign{\vskip 0.5ex}
\nopagebreak Bioreactor & $-1.39/{-1.39}$ & NA & 10 & 1 \\[2pt]
\hline\noalign{\vskip 0.5ex}
\nopagebreak Cancer Treatment & 20.57/20.57 & NA & 10 & 1 \\[2pt]
\hline\noalign{\vskip 0.5ex}
\nopagebreak Cart-Pole Swing-Up & 87.78/242.25 & $\begin{bmatrix} 0/{-0.95} \\ 0/{0.23} \\ 0/{-0.85} \\ 0/{6.63} \end{bmatrix}$ & 100 & 7 \\[8pt]
\hline\noalign{\vskip 0.5ex}
\nopagebreak Glucose & 1354.02/1354.02 & NA & 10 & 1 \\[2pt]
\hline\noalign{\vskip 0.5ex}
\nopagebreak Mould Fungicide & 23.50/23.50 & NA & 10 & 1 \\[2pt]
\hline\noalign{\vskip 0.5ex}
\nopagebreak Mountain Car & 8.57/15.87 & $\begin{bmatrix}0/0.03 \\ 0/{-0.01}\end{bmatrix}$ & 100 & 5 \\[5pt]
\hline\noalign{\vskip 0.5ex}
\nopagebreak Pendulum & 25.42/20.88 & $\begin{bmatrix}-0.01/0.37 \\ -0.01/0.61\end{bmatrix}$ & 100 & 1 \\[5pt]
\hline\noalign{\vskip 0.5ex}
\nopagebreak Predator Prey & 1.79/1.95 & 0/0.14 & 100 & 5 \\[2pt]
\hline\noalign{\vskip 0.5ex}
\nopagebreak Timber Harvest & $-5104.67/{-3928.77}$ & NA & 100 & 1 \\[2pt]
\hline\noalign{\vskip 0.5ex}
\nopagebreak Tumour & 7571.67/8468.68 & NA & 100 & 2 \\[2pt]
\hline\noalign{\vskip 0.5ex}
\nopagebreak Van Der Pol & 2.87/11.12 & $\begin{bmatrix} 0/{-0.35} \\ 0/{-1.76} \end{bmatrix}$ & 10 & 3 \\[3pt]
\bottomrule \\
\end{tabular}
\caption[Learned Neural ODE model performance across variety of systems]{
Summary of performance of trajectory optimization on Neural ODE models learned from data on a variety of environments. 
The layout is the same as the previous table, except that 
parameters are omitted, since there is no way to directly compare
the true parameters with the weights and biases of the 
neural network. Instead, the fourth column shows
the number of trajectories per dataset used in training,
and the fifth column shows the total number
of datasets used to train the model.}
\label{tab:node_mle_sysid}
\end{table}

\clearpage
\section{End-to-end algorithm performance}
\label{app:end_to_end_performance}

In Table~\ref{tab:node_e2e_sysid} we present the performance of running the end-to-end implicit planning imitation learning algorithm described in Section~\ref{sec:end-to-end}, for 13 of the 17 real-world environments, and for the ``simple case'' environment.
To reproduce these results, choose a system in \texttt{config.py} and uncomment \texttt{run\_node\_endtoend} in \texttt{run.py}.
Use default hyperparameters, except as indicated in the final table column, which shows the size of the two hidden layers of the Neural ODE.
To choose a hidden layer size, we tried running smaller (50, 50) and larger (100, 100) networks, and selected the one that achieved better performance on the given environment.
One other important hyperparameter is the number of implicit planning steps to perform at each iteration. 
We found 5 steps (default in the code; found via manual search) to strike a good balance in providing a planning inductive bias while still enabling fast propagation of gradients.

\setlength\LTleft{0pt}
\setlength\LTright{0pt}
\begin{table}[H]
\centering
\begin{tabular}{@{}l l l l@{}}
\toprule
System & Cost (best/achieved) & Defect (best/achieved) & Neurons/Layer \\
\specialrule{.3em}{.2em}{.2em}
\nopagebreak Bacteria & $-7.98$/$-7.60$ & NA & 100 \\[2pt]
\hline\noalign{\vskip 0.5ex}
\nopagebreak Bear Populations & $12.28$/$12.40$ & NA & 100 \\[2pt]
\hline\noalign{\vskip 0.5ex}
\nopagebreak Bioreactor & $-1.39$/$-1.39$ & NA & 100 \\[2pt]
\hline\noalign{\vskip 0.5ex}
\nopagebreak Cancer Treatment & 20.57/20.58 & NA & 50 \\[2pt]
\hline\noalign{\vskip 0.5ex}
\nopagebreak Cart-Pole Swing-Up & 87.78/36.90 & $\begin{bmatrix} 0/1.21 \\ 0/{-4.88} \\ 0/{-0.29} \\ 0/{-3.28} \end{bmatrix}$ & 100 \\[5pt]
\hline\noalign{\vskip 0.5ex}
\nopagebreak Glucose & 1354.02/1354.12 & NA & 100 \\[2pt]
\hline\noalign{\vskip 0.5ex}
\nopagebreak HIV Treatment & $-823.13$/$-822.96$ & NA & 100 \\[2pt]
\hline\noalign{\vskip 0.5ex}
\nopagebreak Mould Fungicide & 23.50/23.54 & NA & 50 \\[2pt]
\hline\noalign{\vskip 0.5ex}
\nopagebreak Mountain Car & 8.57/3000 & $\begin{bmatrix}0/{-0.39} \\ 0/{-0.02} \end{bmatrix}$ & 50 \\[5pt]
\hline\noalign{\vskip 0.5ex}
\nopagebreak Pendulum & 25.53/1.90 & $\begin{bmatrix}0/{-2.69} \\ 0/{-0.08}\end{bmatrix}$ & 100 \\[5pt]
\hline\noalign{\vskip 0.5ex}
\nopagebreak Predator Prey & 1.79/1.04 & 0/${-1.84}$ & 100 \\[2pt]
\hline\noalign{\vskip 0.5ex}
\nopagebreak Timber Harvest & $-5104.67$/$-500.00$ & NA & 50 \\[2pt]
\hline\noalign{\vskip 0.5ex}
\nopagebreak Tumour & 7571.67/9161.23 & NA & 50 \\[2pt]
\hline\noalign{\vskip 0.5ex}
\nopagebreak Van Der Pol & $2.87$/$16.81$ & $\begin{bmatrix} 0/{0.54} \\ 0/{-1.75} \end{bmatrix}$ & 100 \\[3pt]
\bottomrule \\
\end{tabular}
\caption{
Summary of performance of end-to-end learning and planning with Neural ODE models on a variety of environments. 
The first column lists the environment.
The second column indicates the cost of applying the controls solved for with the model, applied
in the true environment.
The third column shows the defect of the final state from the desired final state, if any.
The fourth column shows the size of the hidden layers of the neural network that was used for the model.}
\label{tab:node_e2e_sysid}
\end{table}

\clearpage

\section{Description of the direct multiple shooting algorithm}

The direct single shooting approach does not allow us to impose constraints on the state trajectory, and is inherently sequential.
Direct multiple shooting addresses both these shortcomings by breaking the problem into a sequence of \emph{shooting intervals} on which direct single shooting can be applied in parallel.
As such, it is sometimes advantageous to employ direct multiple shooting instead of direct single shooting.

\begin{figure}[H]
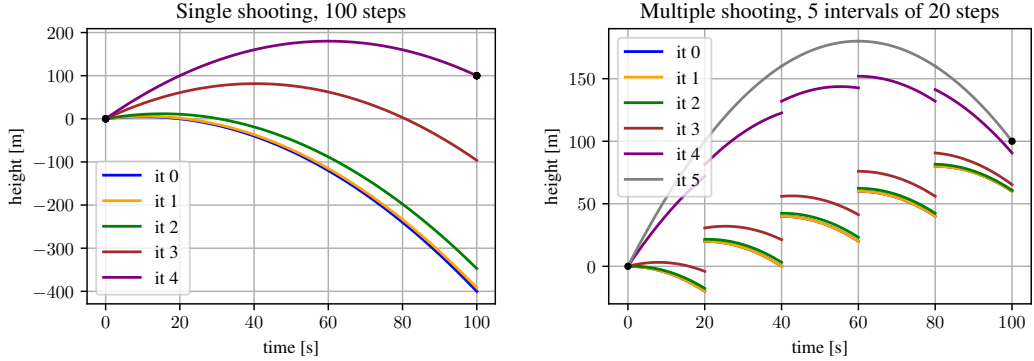

    \centering
    \begin{subfigure}[b]{0.49\textwidth}
        \centering
        \resizebox{\textwidth}{!}{\input{figures/ss.pgf}}
        \caption{Trajectories resulting from 0 to 4 iterations of
        direct single shooting.}
        \label{fig:app_ss}
    \end{subfigure}
    \hfill
    \begin{subfigure}[b]{0.49\textwidth}
        \centering
        \resizebox{\textwidth}{!}{\input{figures/ms.pgf}}
        \caption{Trajectories resulting from
        0 to 5 iterations of direct multiple shooting.}
        \label{fig:app_ms}
    \end{subfigure}
    \caption[Comparison of single and multiple shooting]{
    Comparison of direct single shooting (a) and direct multiple shooting (b), applied to a toy dynamics problem.
    While direct single shooting converges in fewer iterations, each iteration takes longer than those performed in direct multiple shooting.
    Additionally, the parallelism of direct multiple shooting might enable it to tackle problems with longer time horizons.
    }
    \label{fig:app_ss_vs_ms}
\end{figure}

The nonlinear program resulting from direct multiple shooting is presented below, followed by a comparison of the two direct shooting techniques in Figure~\ref{fig:app_ss_vs_ms}, on the same toy problem presented in Section~\ref{sec:traj_opt}.
\begingroup
\allowdisplaybreaks
\begin{equation}
\begin{aligned}
    \text{decision variables} \quad & \hat \x_0,
    \hat \x_k, \hat \x_{2k}, \ldots, \hat \x_{N-k}, \hat \x_N, 
    \hat \u_0, \hat \u_1, \hat \u_2, \ldots, \hat \u_N \\
    \text{objective} \quad & \left[ \sum_{j=1}^{N/k} \int_{t_{(j-1)k}}^{t_{jk}} \f\left( \begin{bmatrix} \x(t) \\ c(t) \end{bmatrix}, \psi(\hat \u_{(j-1)k:jk}, t), t \right) \; dt \right] \texttt{[-1]} \\
    \text{equality constraints} \quad  
    & \hat \x_{ik} = \hat \x_{(i-1)k} +
    \int_{t_{(i-1)k}}^{t_{ik}} f(\x(t), \psi(\hat \u_{(i-1)k:ik}, t)) \quad \text{for } i = 1, 2, \ldots, N/k \\
    & \hat \x_0 = \x_s \\
    * \quad & \hat \x_N = \x_f\\
    \text{inequality constraints*} \quad & \x^\text{lower}_{ik} \leq \hat \x_{ik} \leq \x^\text{upper}_{ik} \quad \text{for } i = 0, 1, \ldots, N/k \\
    * \quad & \u^\text{lower}_{i} \leq \hat \u_{i} \leq \u^\text{upper}_{i} \quad \text{for } i = 0, 1, \ldots, N \\
\end{aligned}
\label{eq:ms}
\end{equation}
\endgroup

\clearpage

\section{Incorporating inequality constraints}
\label{app:inequality_constraints}

For real-world systems, we usually would like to restrict the agent to only take certain paths through state space, avoiding dangerous or otherwise undesirable areas.
Such restrictions can be expressed as bounds on the state variables in the nonlinear program resulting from transcription of the trajectory optimization problem.
Note that in the single shooting setting, there are no state decision variables; this technique is only possible when using multiple shooting or collocation techniques.

Once we have inequality constraints in the nonlinear program, we must find a solution which satisfies these constraints.
When using a Lagrangian-based approach, there are at least two straightforward ways of doing this.
The first approach, which is implemented in Myriad, is \emph{projection}.
Over the course of optimization, after a gradient step is taken, the resulting iterate is projected back into the feasible set. 
With fixed bounds, this can be implemented as a \texttt{clip} operation \citep{bertsekas}.

An alternative approach, which can be included directly in the system definition, is that of \emph{reparametrization}.
Instead of stepping and then modifying the iterate to satisfy the bounds, we instead modify the space in which we are performing the optimization, so that any point in the space will be feasible \citep{inequality}.
For example, if our feasible set is $x_\text{lower} \leq x \leq x_\text{upper}$, a viable reparametrization would be to use a sigmoid of the form $\sigma(x, x_\text{lower}, x_\text{upper}) = (x_\text{upper} - x_\text{lower})/(1 + e^{-\alpha x}) - x_\text{lower}$, where $\alpha$ is a temperature constant which can be decreased over time.

\section{Compute}

All experiments were performed on a personal laptop with the following specifications:
\begin{itemize}
    \item 2.7 GHz Quad-Core Inter Core i7
    \item Intel Iris Plus Graphics 665 1536 MB
    \item 16 GB 2133 MHz LPDDR3
    \item 500 GB PCI-Express SSD
\end{itemize}

The average runtime for experiments is presented in Table \ref{tab:runtimes}.

\begin{table}[H]
\centering
\begin{tabular}{@{}ll@{}}
\toprule
Experiment    & Runtime \\ 
\specialrule{.3em}{.2em}{.2em}
Trajectory Optimization & $\sim 1$ minute per environment \\
\hline\noalign{\vskip 0.5ex}
System Identification (Parametrized) & $\sim 2$ minutes per environment \\
\hline\noalign{\vskip 0.5ex}
System Identification (Neural ODE) & $\sim 2$ hours per environment \\
\hline\noalign{\vskip 0.5ex}
End-to-end Control (Parametrized) & $\sim 0.5$ hours per environment \\
\hline\noalign{\vskip 0.5ex}
End-to-end Control (Neural ODE) & $\sim 10$ hours per environment \\
\hline\noalign{\vskip 0.5ex}
Total for all experiments (above times 18) & $\sim 300$ hours \\
\bottomrule
\\[-5pt]
\end{tabular}
\caption{The approximate amount of compute used for all experiments.}
\label{tab:runtimes}
\end{table}

\end{document}